\documentclass{article}
\usepackage{floatrow}
\newfloatcommand{capbtabbox}{table}[][\FBwidth]

\usepackage[preprint]{nips_2018}

\usepackage[utf8]{inputenc} 
\usepackage[T1]{fontenc}    
\usepackage{hyperref}       
\usepackage{url}            
\usepackage{booktabs}       
\usepackage{amsfonts}       
\usepackage{nicefrac}       
\usepackage{microtype}      
\usepackage{graphicx}
\usepackage{subfigure}
\usepackage{multirow}
\usepackage[utf8]{inputenc}
\usepackage[small]{caption}
\usepackage{CJK}
\usepackage{amsmath}
\usepackage{natbib}
\title{A Novel Framework for Recurrent Neural Networks with Enhancing Information Processing and Transmission between Units}

\author{
Xi Chen,
Zhihong Deng,
Gehui Shen,
Ting Huang
\\
School of Electronics Engineering and Computer Science, Peking University \\
}

\begin{document}

\maketitle

\begin{abstract}
    This paper proposes a novel framework for recurrent neural networks (RNNs) inspired by the human memory models in the field of cognitive neuroscience to enhance information processing and transmission between adjacent RNNs’ units. The proposed framework for RNNs consists of three stages that is working memory, forget, and long-term store. The first stage includes taking input data into sensory memory and transferring it to working memory for preliminary treatment. And the second stage mainly focuses on proactively forgetting the secondary information rather than the primary in the working memory. And finally, we get the long-term store normally using some kind of RNN’s unit. Our framework, which is generalized and simple, is evaluated on 6 datasets which fall into 3 different tasks, corresponding to text classification, image classification and language modelling. Experiments reveal that our framework can obviously improve the performance of traditional recurrent neural networks. And exploratory task shows the ability of our framework of correctly forgetting the secondary information.
\end{abstract}

\section{Introduction}

In recent years, recurrent neural networks have become more widely-used in various tasks, and thus much work has been done on improving conventional recurrent neural network \cite{4} or its improved versions \cite{arv,a1,r1,grd,trr}. However, most of the work just focuses on improving the inner structure of a certain RNNs’ unit or building a model that only based on a certain RNNs’ unit. And there exists little work on improving the way of information processing and transmission between adjacent RNNs’ units.

As we know, in the field of cognitive neuroscience, information processing and transmission is the key component of human memory.  Inspired by the work on human memory, we want to introduce similar mechanics to handle the problem of information processing and transmission in RNNs in this paper.

In the field of cognitive neuroscience, most researchers think that information processing can be divided into relatively independent stages in the process of learning and memory based on many experimental results. They thus put forward many memory models. Among them, modal model \cite{0} proposed by Richard Atkinson and Richard Shiffrin is the most well-known. Modal model holds the view that sensory information is firstly detected by the senses called sensory registers, and then transferred into short-term store. Once the information get into short-term store, it can further enter long-term store when the information is rehearsed. What’s more, the model also thinks that information can be lost at every stage because of decline (information loss and gradual disappearance), interference (new information replacing the old), or the combination of them.

Since modal model was proposed, intense debates on the model have continued. Some key questions can be summarized as 1) whether information should be encoded in short-term memory before it can be stored in long term memory, 2) does short-term memory just store the information. A lot of researches have been done on the aforementioned questions. For the first question, some work such as \cite{wwww} supports the view that information can be directly stored in long-term memory without passing through the short-term memory. And for the second question, researchers like Baddeley pointed out that short-term memory is not enough to explain that the information can be processed in a very short time, and thus put forward the working memory system to improve the short-term store \cite{bw}. The working memory proposed emphasizes that the information not only can be temporarily stored in it, but also can be attended to and manipulated.

The research on human memory is still going on, but in this paper we only focus on the classical models like modal model and some developed models or theories such as Baddeley's model of working memory for simplicity and applicability. We will introduce the main theories on human memory to build a novel framework for RNNs.

In respect of the aspect about enhancing the information processing and transmission in RNNs, we aim at making our framework has ability of manually controlling the process of “forgetting” to help decline more secondary information so that new information prefers to replace more secondary old information rather than the primary. Our “forgetting” mechanism is different from what in the long short-term memory neural network(LSTM) \cite{4} or its improved versions of it, which we will describe in more detail in the later sections.

And the other important goal is to try to make our framework be compatible with all RNNs’ units. It’s nearly not be taken into account in previous related work, such as \cite{arv}, \cite{a1}, \cite{r1}, \cite{grd} and \cite{trr}. The previous work pays close attention to building a model that only based on a certain RNNs’ unit like LSTM unit or improving the inner structure of a certain RNNs’ unit, and thus can’t be easily transferred to other RNN models that based on other RNNs’ units. Instead, the major characteristic of our work lie in the way of processing and transmission of information between adjacent RNNs' units.

Simply put, the contributions of this paper can be summarized as follows.

\begin{itemize}
\item As far as we know, our work is the first one that introduces a relatively complete human memory model (modal model but modified based on later models or theories) in RNNs to enhance the information processing and transmission.
\item A specific recurrent neural network framework is proposed, which is generalized and simple.
\item The experimental results on both text and image classification tasks and language modelling task reveal that our framework can obviously improve the performance of traditional recurrent neural networks.
\item We test and verify our framework’s ability of correctly forgetting the secondary information through exploratory experiments.
\end{itemize}

The outline of the rest of the paper is as follows. We first, in Section 2, give a brief background on modal model, Baddeley's model of working memory and RNNs. We then propose our recurrent neural network framework in Section 3. Section 4 shows our experiments, which include text and image classification tasks, a language modelling task and an exploratory task. Section 5 concludes the whole paper.

\section{Background}

\subsection{Modal Model and Baddeley's Model of Working Memory}
Modal model first described by Atkinson and Shiffrin in 1968 \cite{0} is a human memory model that explains how human memory processes work. The model consists of three separate components: the sensory register, short-term store, and long-term store.

The sensory register’s role is to detect the environmental stimulus and hold them for use in short-term store. The sensory registers is like "buffers" which doesn’t process the information. And the information decays rapidly and is forgotten when it’s not transferred to the short-term store.

Short-term store receives and holds input from the sensory register, and it can also extract the information from the long-term store. And as with sensory memory, though the information in the short-term store can be held for much longer, it decays and is lost. But when the information transfers to the long-term store through rehearsal, it can more or less be stored permanently.

The long-term store receives the information only from the short-term store, but the information in long-term store can be transferred to the working memory where it can be manipulated.

Modal model is surely not that perfect, and there are some controversial points which we have mentioned in Section 1. But as the beginning of the study, it can also be the basis of our framework for its simplicity and applicability.

Here we in passing give a brief introduction to Baddeley's model of working memory. The early model of working memory is put forward by Alan Baddeley and Graham Hitch in 1974 \cite{bw}. And there is also a lot of follow-up work such as \cite{aq} and \cite{rf}. Baddeley's model of working memory proposed three-part working memory as an alternative to the short-term store in modal model. The three parts include visuo-spatial working memory, phonological loop and central executive. They think the information in working memory is not only just stored, but also processed temporarily, which is different from the definition of the short-term store. We’ll simply absorb this viewpoint in our framework rather than specifically imitating the three parts.

\subsection{Recurrent Neural Networks}

Recurrent neural networks is an extension of a conventional feedforward neural network, which is well adapted for handling a variable-length sequence input. It has various applications in different fields such as natural language processing \cite{ali,ba}, speech processing \cite{gr}, image processing \cite{db,co} and video processing \cite{ve}.
The RNNs ordinarily receive a sequence $x = (x_1, x_2, \ldots, x_T)$ as input, where $x_t = R^d$. And it updates its recurrent hidden state $h_t$ by
{\setlength\abovedisplayskip{5pt}
\setlength\belowdisplayskip{5pt}
\begin{equation}\label{1}
    h_t = \varphi(h_{t-1}, x_t)
\end{equation}}
where $\varphi$ is a smooth, bounded function. And traditionally, the update of the recurrent hidden state in Eq.(1) is implemented as:
{\setlength\abovedisplayskip{5pt}
\setlength\belowdisplayskip{5pt}
\begin{equation}\label{2}
    h_t = \sigma (Wx_t + Uh_{t-1} + b)
\end{equation}}
where $\sigma$ is the logistic sigmoid function, \emph{W} and \emph{U} are parameter matrices, \emph{b} is bias.

The long short-term memory neural network(LSTM) \cite{4} is one of the most important extension of the recurrent neural network. It has an interesting and special implementation of Eq.(1), which is given below:
{\setlength\abovedisplayskip{5pt}
\setlength\belowdisplayskip{5pt}
\begin{flalign}
 &i_{t} = \sigma(W_{ix}x_{t} + W_{ih}h_{t-1} + W_{ic}c_{t-1})  \label{3} \\
 &f_t = \sigma(W_{fx}x_t + W_{fh}h_{t-1} + W_{fc}c_{t-1})  \label{4} \\
 &c_t = f_t \odot c_{t-1} + i_t \odot \phi(W_{cx}x_t + W_{ch}h_{t-1})  \label{5} \\
 &o_{t} = \sigma (W_{ox}x_{t} + W_{oh}h_{t-1} + W_{oc}c_t)  \label{6} \\
 &h_{t} = o_t \odot \phi(c_t)  \label{7}
\end{flalign}
}
That repeating module is called LSTM cell, which consists of a memory cell \emph{c}, an input gate \emph{i}, a forget gate \emph{f}, an output gate \emph{o}. And $\sigma$ and $\phi$ are the logistic sigmoid function and hyperbolic tangent function respectively; $i_t$, $f_t$, $o_t$ and $c_t$ are respectively the input gate, forget gate, output gate, and memory cell activation vector at time step \emph{t}. \emph{$W_{..}$} denotes corresponding parameter matrice. $\odot$ denotes the element-wise multiplication operator.

\section{The Proposed Framework}

\begin{figure*}
 \centering
 \subfigure[Conventional framework for RNNs]{
 \label{fig2:subfig2:a} 
 \includegraphics[width=2.5in]{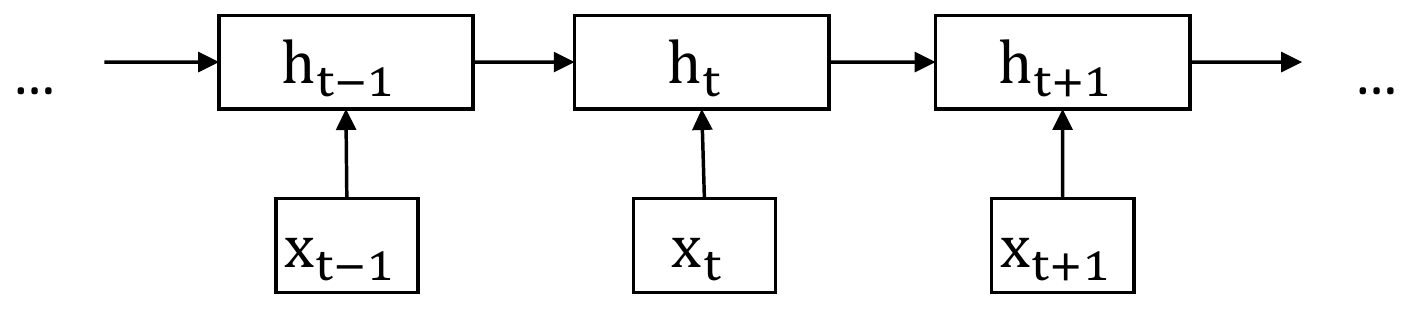}}
 \hspace{0.3in}
 \subfigure[The framework we proposed]{ \label{fig2:subfig2:b} 
 \includegraphics[width=2.5in]{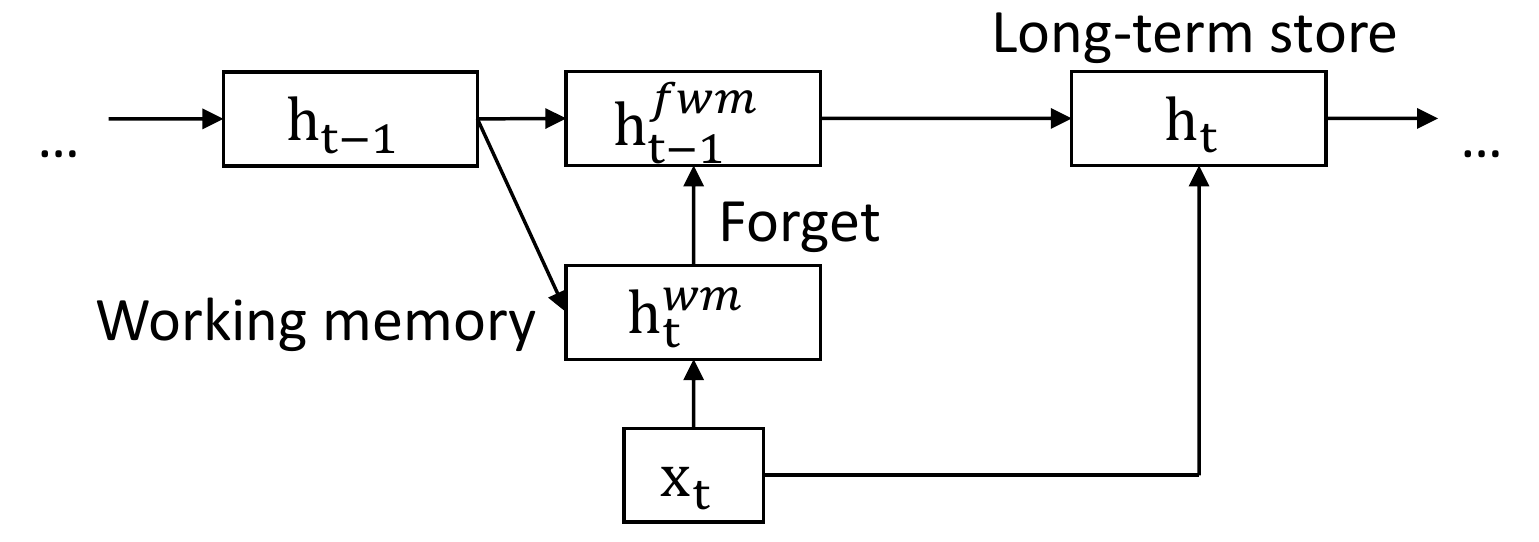}}
 \caption{Comparision between conventional RNNs' framework and the framework we proposed.}
 \label{fig2:subfig2} 
 \end{figure*}

\subsection{Architecture}
Our proposed RNNs framework consists of three stages: working memory, forget, and long-term memory.
The definitions of all these three stages are as follows:
\subsubsection{Working Memory}
This stage includes taking input data into sensory memory and transferring it to working memory for preliminary treatment. We implement it by:
{\setlength\abovedisplayskip{5pt}
\setlength\belowdisplayskip{5pt}
\begin{equation}\label{8}
    h_{t}^{wm} = g(W_{wm}x_t + U_{wm}h_{t-1} + b_{wm})
\end{equation}
}
where \emph{g} is a nonlinear function, $W_{wm}$ and $U_{wm}$ are parameter matrices, $b_{wm}$ is bias. $h_{t}^{wm}$ denotes the \emph{t}-th working memory. The calculation is the same as what we use to calculate the \emph{t}-th recurrent hidden state in conventional RNN.
The reason why we use Eq.(8) is that: $x_t$ and $h_{t-1}$ can be seen as the \emph{t}-th sensory information and (\emph{t}-1)th long-term store respectively. And using the sensory information and (\emph{t}-1)th long-term store $h_{t-1}$ to calculate the \emph{t}-th working memory satisfies two important theories we have mentioned that 1) working memory can receive and hold input from the sensory register, and 2) the information in long-term store can be transferred to the working memory. Besides, the calculation of Eq. (8) is simple and common, which is of equal importance.

\subsubsection{Forget}

Baddeley's model believes that information not only can be temporarily stored in the working memory, but also can be manipulated. Also, models including modal model hold the view that information can be lost at every stage because of decline, interference, or the combination of them. And since the lost is hard to avoid, we mainly focus on how to manually control the process in the working memory. In another word, we are trying to proactively forget the secondary information rather than the primary in the working memory at this stage. We implement it as:
{\setlength\abovedisplayskip{5pt}
\setlength\belowdisplayskip{5pt}
\begin{flalign}
 &F_t = f(h_t^{wm}, h_{t-1})  \label{9} \\
 &h_{t-1}^{fwm} = F_t \odot h_{t-1}  \label{10}
\end{flalign}
}
where $F_t$ can be seen as a \emph{forget weight} vector, and thus \emph{f} denotes a function, which is used to compute the \emph{forget weight} for the (\emph{t}-1)th long-term store. The function can be a variety of forms. We use the following two forms in our paper:
{\setlength\abovedisplayskip{5pt}
\setlength\belowdisplayskip{5pt}
\begin{flalign}
 &f^1 = \sigma(W_f h_t^{wm} + b_f)  \label{11} \\
 &f^2 = \sigma(h_t^{wm} \odot h_{t-1})  \label{12}
\end{flalign}
}
The first form simply computes a weighted sum of the working memory and applies sigmoid function to limit the output between 0 and 1. It doesn’t need to utilize the (\emph{t}-1)-th long-term store $h_{t-1}$. The parameter matrices $W_f$ and $b_f$ are shared throughout the time steps.
The second form do a dot products between the working memory $h_t^{wm}$ and the (\emph{t}-1)-th long-term store $h_{t-1}$. Each dimension of the output indicates a similarity between corresponding dimension of $h_t^{wm}$ and $h_{t-1}$. The smaller the value of a certain dimension is, the larger impact caused by $x_t$ on this dimension is, which means we should forget more information in this dimension on $h_{t-1}$. Note that we do not need to introduce extra parameters in this form.

\subsubsection{Long-term Store}
This stage computes the \emph{t}-th long-term store $h_t$. We implement it as:
{\setlength\abovedisplayskip{5pt}
\setlength\belowdisplayskip{5pt}
\begin{equation}\label{13}
    h_t = {func}_{rnn}(x_t, h_{t-1}^{fwm})
\end{equation}
}
where ${func}_{rnn}$ can be any of RNNs’ units, such as basic RNN unit, LSTM unit, GRU or some other forms like \cite{arv}. It reflects the extendibility of our framework. Take basic RNN unit and LSTM unit as examples.
For basic RNN unit, we update $h_t$ by:
{\setlength\abovedisplayskip{5pt}
\setlength\belowdisplayskip{5pt}
\begin{equation}\label{14}
    h_t = g (W_{ls}x_t + U_{ls}h_{t-1}^f + b_{ls})
\end{equation}
}
where $W_{ls}$, $U_{ls}$ and $b_{ls}$ share with the corresponding parameters at the stage of the working memory.

For LSTM unit, we update $h_t$ by
{\setlength\abovedisplayskip{5pt}
\setlength\belowdisplayskip{5pt}
\begin{flalign}
 &i_{t} = \sigma(W_{ix}x_{t} + W_{ih}h_{t-1}^{fwm} + W_{ic}c_{t-1})  \label{15} \\
 &f_t = \sigma(W_{fx}x_t + W_{fh}h_{t-1}^{fwm} + W_{fc}c_{t-1})  \label{16} \\
 &c_t = f_t \odot c_{t-1} + i_t \odot \phi(W_{cx}x_t + W_{ch}h_{t-1}^{fwm})  \label{17} \\
 &o_{t} = \sigma (W_{ox}x_{t} + W_{oh}h_{t-1} + W_{oc}c_t)  \label{18} \\
 &h_{t} = o_t \odot \phi(c_t)  \label{19}
\end{flalign}
}
They are almost the same as what in LSTM unit. We only replace the conventional LSTM unit’s input $h_{t-1}$ with the previous stage’s output $h_{t-1}^{fwm}$. Note that it is unnecessary to modify the internal structure of the LSTM unit.

\subsection{Discussion}
To further elucidate the characteristics of our framework, in this section, we will analyze our framework from two perspectives and thus distinguish our framework from other work like \cite{4} and \cite{a1}.

Firstly, from the human memory models’ perspective. Comparing the implementation between our framework and previous work like LSTM unit, the most important difference is that LSTM unit doesn’t have the working memory stage. It gets \emph{forget weight} directly using input data and previous long-term store. One of the effect is that LSTM unit can’t forget the information that has stored before. In the most extreme case where we want to discard all the information that has stored before in the \emph{t}-th step and recalculate the state by $x_t$ only, we can’t make $c_{t}$ be independent of both $c_{t-1}$ and $h_{t-1}$. Note that $h_{t-1}$ has been calculated already at the \emph{t}-th step, thus we can’t discard it by set output gate zero.

Secondly, we can also analyze our framework in a more intuitive way. It can be found that in previous work which also introduces the process of “forgetting”, they always forget something directly based on the previous states when processing the sequence. However, it’s more reasonable that the model decides whether or not to forget something after a simple manipulation of the data at the \emph{t}-th step. Here we call that mechanism “look forward and forget”. It means before we have simply processed the data at the \emph{t}-th step, we shouldn’t rashly forget something. Thus, in our framework we can see, we first look forward at the \emph{t}-th step (Eq. (8)), and based on that preliminary recurrent state at the \emph{t}-th step $h_{t}^{wm}$, we calculate the \emph{forget weight} and then forget the information in $h_{t-1}$. After we get new recurrent state $h_{t-1}^{fwm}$, we can then process the data normally using basic RNN unit, LSTM unit or any other RNNs’ units.

\section{Experiments}
We evaluate our framework by applying it to 6 different datasets: MR, CR, TREC, Reuters, MNIST and PTB. These datasets fall into 3 different tasks, corresponding to text classification, image classification and language modelling. Then we also perform an exploratory experiment to validate the ability of our framework of correctly forgetting the secondary information.

\subsection{Datasets}
The datasets for text classification includes MR, CR, TREC and Reuters. The dataset for image classification is MNIST. And Language modelling task evaluates on PTB. Each dataset is briefly described as follows.
\begin{itemize}
\item \textbf{MR:} Movie reviews with one sentence per review \cite{6}. Classification involves detecting positive/negative reviews.
\item \textbf{CR:} Customer reviews of 14 products obtained from Amazon \cite{7}. The goal is to predict positive/negative reviews.
\item \textbf{TREC:} TREC question dataset \cite{trec}, in which the objective is to classify each question into 6 question types.
\item \textbf{Reuters:} Reuters newswire 46 topics classification. A collection of documents that appeared on Reuters newswire in 1987.\footnote{the dataset is available at https://s3.amazonaws.com/text-datasets/reuters.npz}
\item \textbf{MNIST:} Dataset of grayscale images of the 10 digits. The images are centered and of size 28 by 28 pixels.\cite{m}
\item \textbf{PTB:} The Penn Tree Bank dataset is made of articles from the Wall Street Journal.\cite{ptb}
\end{itemize}

\subsection{Implement}
For text categorization task, we use 300-d word embeddings pretrained by Glove on MR, CR and TREC, and fine-tune the word embeddings during training to improve the performance. the dimension of recurrent states is also set to 300. While on Reuters, the words are tokenized already, thus the word embeddings are adjust to 128, and randomly sampled from uniform distribution. Sentence representation is finally created by using the average pooling of all hidden states of the RNN. After obtaining the sentence embedding, we apply fully connected layers followed by a softmax non-linear layer that predicts the probability distribution over classes.

For image categorization task, each MNIST image is input in RNN line by line. The representation of the image is finally created by using the final hidden state of the RNNs. And we also apply a fully connected layer followed by a softmax non-linear layer to predict the classes.

For test and image categorization task. Training is done through stochastic gradient descent over shuffled mini-batches with the Adam update rule \cite{9}. And for regularization, we only employ dropout as described in \cite{8} on the recurrent units.

As for language modelling task, we use 500-d word embeddings whose dimension is equal to the recurrent state’s. And other settings and training methods are the same as \cite{zzz}.

\subsection{Results}

\subsubsection{Text Classification}
Table 1 shows the text classification accuracies of the baseline models along with our four models F+RNN, F*+RNN, F+LSTM and F*+LSTM, where RNN and LSTM denote 1-layer unidirectional conventional RNN and 1-layer unidirectional LSTM respectively, F and F* denote that $f^1$ or $f^2$ is used at the stage of forget in our framework. And we will use the same notations next except that in the language modelling task where LSTM denotes 2-layer network.

As we can see, for the models based on the basic RNN unit, both F+RNN and F*+RNN achieve higher accuracies than RNN on all four datasets except F*+RNN on TREC only. And it is worth noting that F+RNN even outperforms LSTM on MR, CR and TREC, and just about 1\% lower on Reuters. Similarly, for the models based on LSTM unit, our F+LSTM and F*+LSTM also surpass LSTM definitely. These results demonstrate the advantage of our framework.

And for our two kinds of the models which has different ways of calculating the forget weight respectively, F+RNN consistently outperformd F*+RNN on all four datasets, and F*+LSTM only achieves higher accuracies than F+LSTM on MR and TREC. We explain it from the perspective of the complexity of two models. Since no extra parameters are introduced in the stage of the forget in F*+RNN and F*+LSTM for the sake of keeping the models light-weight, they calculate the \emph{t}-th step forget weight only on account of the difference between the \emph{t}-th working memory and the (\emph{t}-1)th long-term store, which of course suffers from the lack of flexibility in the majority of cases. Besides, it’s also interesting that F+RNN slightly outperform F+LSTM on MR ,CR and TREC. The results may probably due to the simplicity of the datasets.

\begin{table}
\centering
\begin{tabular}{|p{3cm}|p{2cm}|p{2cm}|p{2cm}|p{2cm}|}
  \hline
  Model & MR & CR & TREC & Reuters \\
  \hline
  RNN & 78.5\% & 81.0\% & 90.5\% & 76.7\% \\
  \hline
  F+RNN & \textbf{80.3\%} & \textbf{82.0\%} & \textbf{92.8\%} & \textbf{78.3\%} \\
  \hline
  F*+RNN & 79.3\% & 81.8\% & 89.8\% & 77.0\% \\
  \hline\hline
  LSTM & 79.0\% & 81.5\% & 90.5\% & 79.5\% \\
  \hline
  F+LSTM & 80.2\% & \textbf{81.9\%} & 91.8\% & \textbf{82.0\%} \\
  \hline
  F*+LSTM & \textbf{80.9\%} & 81.6\% & \textbf{92.2\%} & 81.6\% \\
  \hline
\end{tabular}
  \caption{Results of the text classification accuracies of the baseline models along with our four models F+RNN, F*+RNN, F+LSTM and F*+LSTM.}
\end{table}

\subsubsection{Image Categorization}
Since the models based on the conventional recurrent unit perform poor on MNIST, we only compare the models based on LSTM. The experimental results can be seen in Table 2. Though the accuracy of F*+LSTM is lower than LSTM’s, F+LSTM achieves the higher accuracy. It can be inferred that F*+LSTM doesn't correctly forget the secondary information as we expect on MNIST since a lack of parameters at the stage of forget compared to F*+LSTM. It also suggests that correctly forgetting the secondary information is not always easy.

\begin{table*}
\begin{floatrow}
\capbtabbox{
\begin{tabular}{|p{3cm}|p{3cm}|}
  \hline
  Dataset & MNIST \\
  \hline
  LSTM & 98.0\%\\
  \hline
  F+LSTM & \textbf{98.3\%}\\
  \hline
  F*+LSTM & 97.7\%\\
  \hline
\end{tabular}
}{
 \caption{Results of the image classification accuracies of the baseline models compared with the models in the base of our framework.}
 \label{tab:tb1}
}
\capbtabbox{
\begin{tabular}{|p{3cm}|p{3cm}|}
  \hline
  Dataset & PTB \\
  \hline
  LSTM & 89.335\\
  \hline
  F+LSTM & \textbf{84.899}\\
  \hline
  F*+LSTM & 87.566\\
  \hline
\end{tabular}
}{
 \caption{Test perplexity of language modelling task for LSTM, LSTM+F and LSTM+F* on PTB.}
 \label{tab:tb2}
}
\end{floatrow}
\end{table*}

\subsubsection{Language Modelling}

In this subsection, we report the test perplexity of language modelling task for F+LSTM, F*+LSTM and LSTM on PTB. In Table 3, we observe that our F+LSTM and F*LSTM both achieve lower perplexity than LSTM definitely. And LSTM+F performs better than F*+LSTM by reducing about 3 perplexity.

\subsection{Exploratory Experiments}

\begin{figure*}
 \centering
 \subfigure[training set]{
 \label{fig2:subfig2:a} 
 \includegraphics[width=2.5in]{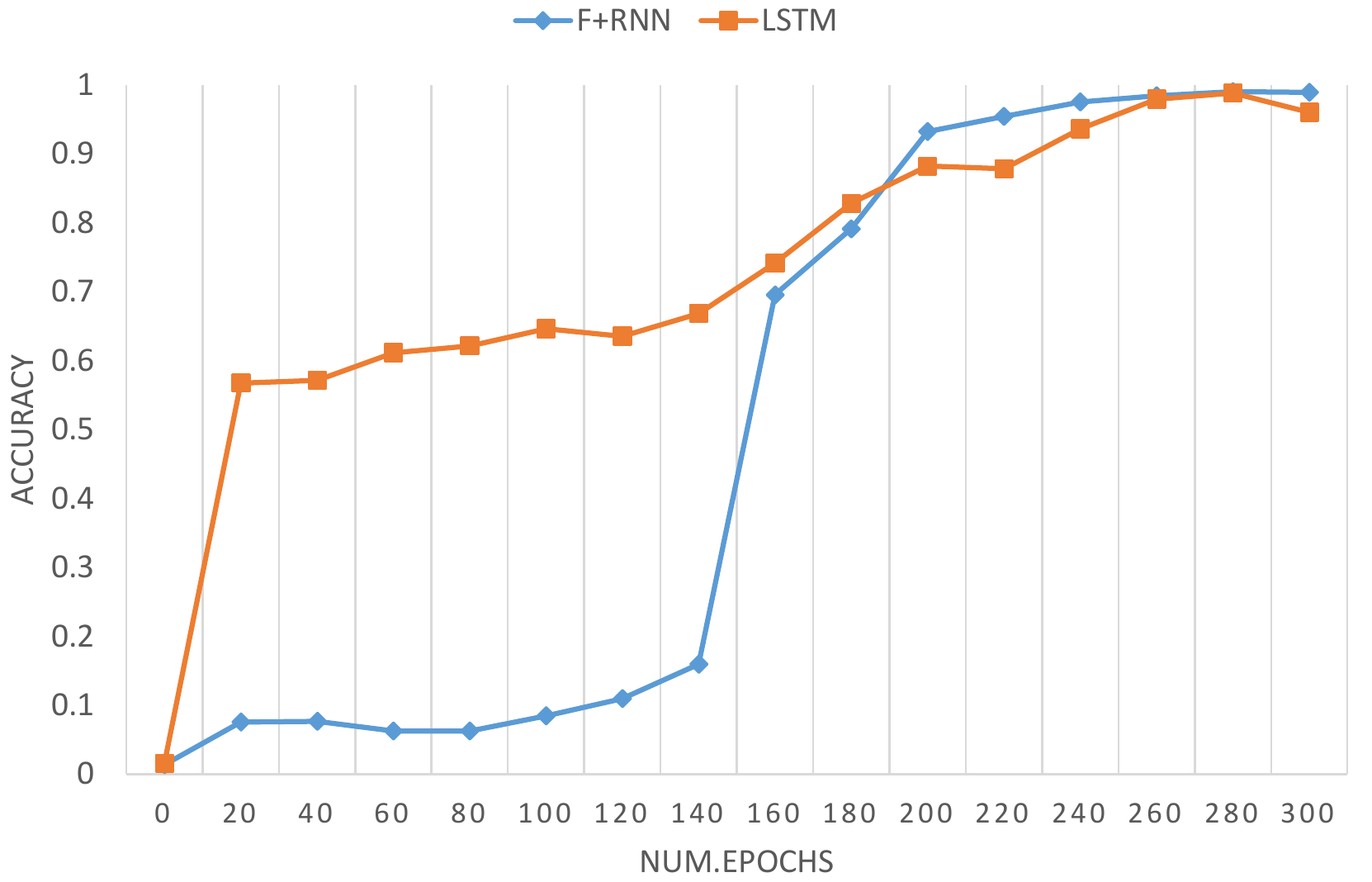}}
 \hspace{0.1in}
 \subfigure[test set]{ \label{fig2:subfig2:b} 
 \includegraphics[width=2.5in]{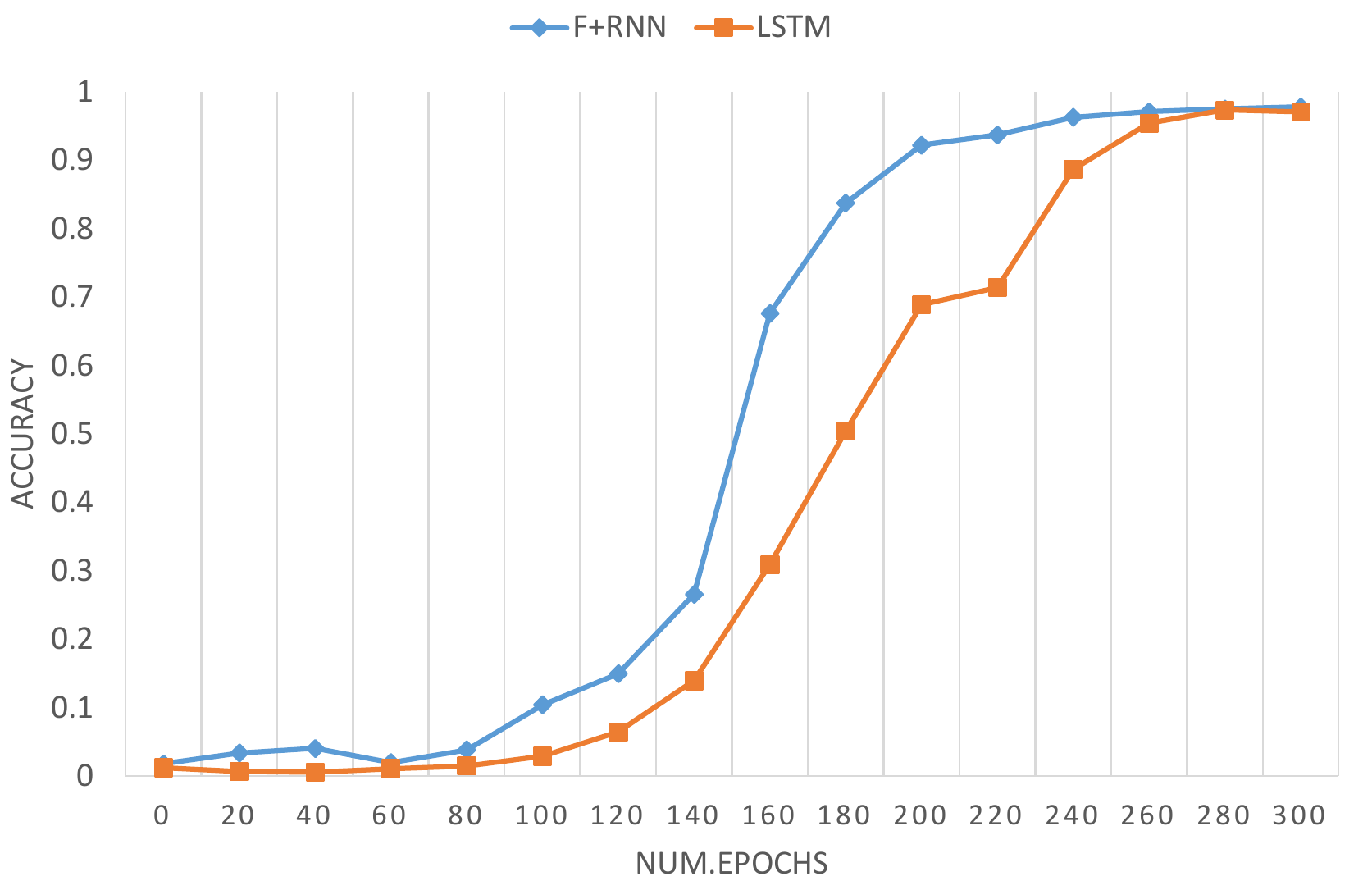}}
 \caption{The dynamical changes of the accuracy on both training set and test set}
 \label{fig2:subfig2} 
 \end{figure*}

 \begin{table*}
\centering
\begin{tabular}{|p{7cm}|p{4cm}|}
  \hline
  Type & Forget Rate (Avg.) \\
  \hline
  the positions where 0,1 appears & 0.5291\\
  \hline
 the positions where -1 appears & 0.0809\\
  \hline
\end{tabular}
  \caption{The comparison of the forget rate between two types of positions in the F+RNN model.}
\end{table*}

\begin{figure*}
  \centering
  \includegraphics[width=0.6\textwidth]{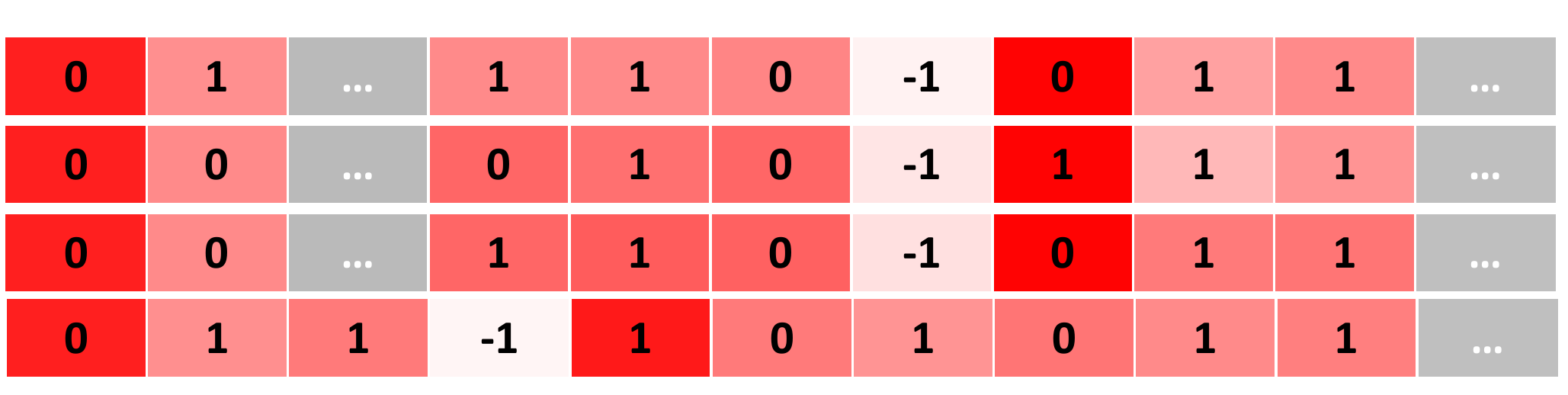}\\
  \caption{Cases of the heat maps for the forget rate}\label{Figure 4}
\end{figure*}

In this subsection we are going to do an exploratory experiment to show the ability of our framework of correctly forgetting the secondary information.

We design a novel dataset that consists of 100,000 digital sequences, each of which is composed of 0, 1 and -1 only and has a length of 100. We restrict the number of -1's in each sequence to a maximum of 1. If -1 appears in a sequence, the output is the number of 1's after the -1. Otherwise, the output is the number of 1's in the whole sequence. And for simplicity, we call the sequence \emph{special} if -1 appears in the sequence. The training set consists of 80,000 sequences in which 30,000 sequences are \emph{special}. While in 20,000 test sequences, all are \emph{special}.

We evaluate the F+RNN model on the dataset compared with LSTM. Note that our F+RNN model is based on the conventional recurrent neural network, which is considered weaker than LSTM in capturing long-term dependencies.

As we can see from Figure 3, which demonstrates the dynamical changes of the accuracy on training set and test set, LSTM only converges faster on the training dataset than F+RNN at the beginning but slower after about 180 epoch, and, most importantly, has a much slower convergence rate on the test dataset. An explanation for the results is that LSTM doesn’t have the working memory like our model, and thus has poor ability to forget the information that has stored before, which we have discussed in 3.2 in detail.

In order to further test the hypothesis, we compare the forget rate at two types of positions on all 20,000 test sequences in our F+RNN model. The first position type denotes the positions where -1 appears, and the second denotes the others. The forget rate at certain position is computed by normalizing the L1-norm of the \emph{forget weight} vector. Note that the forget rate here means the reservation rate of the information that has stored before. As we can see, the forget rate for the first position type is significantly lower than the second type, which demonstrates that our framework has ability of learning when to forget useless information and thus brings efficient information processing.

We also randomly select some sequences from the test dataset and plot the heat maps for the forget rate on these cases (Figure 4). In the heat maps, the greater the forget rate is, the darker the color is. Consistent with the results in the table 4, the forget rates at the positions where -1 appears are much smaller. And the forget rates at the positions after -1 are the greatest in general, which is the same to the position at the beginning of the sequence. It's reasonable since -1 can also be seen as a kind of start symbol in our task.

\section{Conclusion}

In this paper, we introduce a relatively complete human memory model to build a specific recurrent neural network framework to improve the information processing and transmission. The framework manually controls the process of “forgetting” to help decline more secondary information rather than the primary. The framework is generalized, which is compatible with all kinds of recurrent unit including LSTM unit, GRU and even more complex units. Also, the framework benefits from its simplicity, which needs very few extra parameters and is easy to implement.

\bibliographystyle{plainnat}
\bibliography{nips_2018i}

\end{document}